\documentclass[12pt]{article} 

\usepackage{authblk}
\usepackage{times}  
\usepackage{helvet} 
\usepackage{courier}  
\usepackage[hyphens]{url}  
\usepackage{graphicx} 
\usepackage{amsfonts}
\usepackage{amssymb}
\usepackage{amsmath}
\usepackage{multirow}
\usepackage{booktabs}
\usepackage{bm}
\usepackage{graphicx}  
\usepackage{color}
\usepackage{subcaption}
\usepackage{framed}

\usepackage{natbib}
\setcitestyle{authoryear,open={(},close={)}}

\usepackage{hyperref}
 \hypersetup{
     colorlinks=true,
     linkcolor=blue,
     filecolor=blue,
     citecolor = blue,      
     urlcolor = blue,
     }

\title{Living in the Physics and Machine Learning Interplay for Earth Observation}

\author{Gustau Camps-Valls, Daniel H. Svendsen, Jordi Cort\'es-Andr\'es, \'Alvaro Moreno-Mart\'inez, Adri\'an P\'erez-Suay, Jose Adsuara, Irene Mart\'in, Maria Piles, Jordi Mu{\~n}oz-Mar\'i, Luca Martino}
\affil{Image Processing Laboratory (IPL), Universitat de Val\`encia, Spain. http://isp.uv.es -- @isp\_uv\_es}

\date{}

\begin{document}

\maketitle

\begin{abstract}
Most problems in Earth sciences aim to do inferences about the system, where accurate predictions are just a tiny part of the whole problem. Inferences mean understanding variables relations, deriving models that are physically interpretable, that are simple parsimonious, and mathematically tractable. Machine learning models alone are excellent approximators, but very often do not respect the most elementary laws of physics, like mass or energy conservation, so consistency and confidence are compromised. 
In this paper we describe the main challenges ahead in the field, and introduce several ways to live in the Physics and machine learning interplay: to encode differential equations from data, constrain data-driven models with physics-priors and dependence constraints, improve parameterizations, emulate physical models, and blend data-driven and process-based models.
This is a collective long-term AI agenda towards developing and applying algorithms capable of discovering knowledge in the Earth system.
\end{abstract}

\section{Introduction}

Process understanding and modeling is at the core of the scientific reasoning. Principled {\em parametric} and mechanistic modeling has dominated science and engineering until recently with the advent of machine learning. Despite great success in many areas, machine learning algorithms in the Earth sciences face the problem of credibility and consistency, as often they do not respect the most elementary laws of physics like energy or mass conservation. 

Physicists and environmental scientists attempt to model systems in a principled way through analytic descriptions that encode prior beliefs of the underlying processes. Conservation laws, physical principles or phenomenological behaviors are generally formalized using mechanistic models and differential equations. This physical paradigm has been, and still is, the main framework for modeling complex natural phenomena like e.g. those involved in the Earth system. With the availability of large datasets captured with different sensory systems, the physical modeling paradigm is being challenged (and in many cases replaced) by the statistical machine learning (ML) paradigm, which offers a prior-agnostic approach \citep{Reichstein19nat,Halevy09,bezenac2019}.

Machine learning models can fit observations very well, but predictions may be physically inconsistent or even implausible, owing for example to extrapolation or observational biases. This has been perhaps the most important criticism to ML algorithms, and a relevant reason why, historically, physical modeling and machine learning have often been treated as two different fields under very different scientific paradigms (theory-driven versus data-driven). 
Likewise, there is an on-going debate about the limitations of traditional methodological frameworks: both about their scientific insight and discovery limits in general \citep{Halevy09}, and in the geosciences and hydrology in particular \citep{Karpatne17}. 
Recently, however, integration of domain knowledge and achievement of physical consistency by teaching ML models about the governing physical rules of the Earth system has been proposed as a principled way to provide strong theoretical constraints on top of the observational ones \citep{Reichstein19nat}. The synergy between the two approaches has been gaining attention, by either redesigning model's architecture, augmenting the training dataset with simulations, or by including physical constraints in the cost function to minimize  \citep{Karpatne17,svendsen17jgp,Reichstein19nat}. 
Hybrid modeling, however, is more than that, and the concept changes in different communities and disciplines. 

Here we review methodologies originated from the field of Earth observation. We typically rely on model simulations that are costly so emulation comes into play. Such models also need to be inverted with efficient ML algorithms, and their parameters estimated not only point-wise but ideally their distribution should be inferred. In addition, in many Earth and environmental problems, knowledge about the driving/generating process is partly known, described or parameterized: with hybrid modeling one can then learn the latent functions, driving forces as well as their parameters. 
%
The integration of physics in machine learning models may not only achieve improved performance and generalization but, more importantly, may lead to improved consistency and credibility of the machine learning models. As a byproduct, 
Actually, the hybridization has an interesting regularization effect, given that the physics limits the parameter space to search and thus discards implausible models. 
Therefore, physics-aware machine learning models combat overfitting better, become simpler (sparser), and require less amount of training data to achieve similar performance \citep{Stewart16}. 
Physics-aware ML thus lead to enhanced computational efficiency, and constitute a stepping stone towards achieving more interpretable {and robust} machine learning models \citep{Samek2019,von2019informed}. 
We describe the main challenges ahead in the field of Earth observation, and illustrate several ways to live in the Physics and machine learning interplay to address them. Some final conclusions and remarks are drawn.

\section{Context, Challenges and Outline} 

\begin{figure}[h!]
    \centering
    \includegraphics[width=12cm]{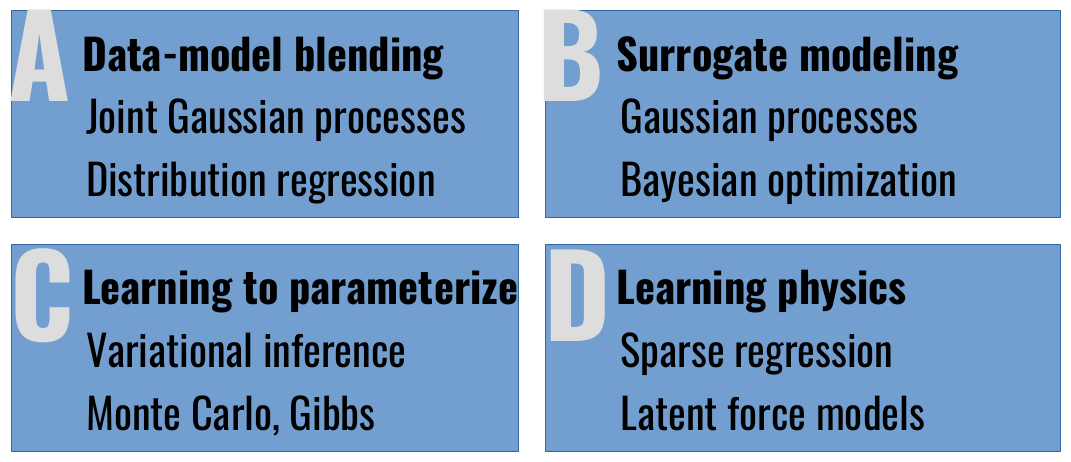}
    \caption{Families and introduced methods to combine physics (mechanistic models, simulations, domain knowledge) and observational data (satellite, airborne, in situ) in Earth observation problems.}   \label{fig:scheme}
\end{figure}

Physics modeling and machine learning have often been considered as completely different and irreconcilable fields in Earth sciences, as in many other disciplines. Yet, these approaches are indeed complementary: physical approaches are interpretable and allow extrapolation beyond the observation space by construction, and data-driven approaches are highly flexible and adaptive to data. Their synergy has gained attention lately in remote sensing and the geosciences~\citep{Karpatne17,camps2018physics}. 
Interactions can be diverse \citep{Reichstein19nat}, but can be grouped in four main overarching goals, for which we provide recent exciting developments, cf. Fig.~\ref{fig:scheme}: (A) blending data-driven and mechanistic (or domain knowledge) models; (B) emulate physical models for efficiency and mathematical tractability; (C) learning parameterizations from data and inverse modeling; and (D) learning physics principles and equations from data.

\subsubsection{A. Blending ML and process-based models} 

ML algorithms and physics can be fully blended in several ways extending traditional approaches based on data assimilation, as suggested in \citep{Reichstein19nat}, but very often simple constrained optimization approaches work well in practice. Including knowledge through extra regularization can be seen as a form of inductive bias \citep{kashinath2019physics,wu2018physics}. Another option is to learn emulators which are then combined with purely data-driven algorithms for model inversion \citep{camps2018physics}. Finally, one can also consider a fully coupled net where layers that describe complicated and uncertain processes feed physics-layers that encode known relations of intermediate feature representation with the target variables. 
The integration of physics into DL models not only allows us to achieve improved generalization properties but, more importantly, endorses DL models with physical consistency. In turn, the hybridization process has an interesting regularization effect as physics discards implausible models and promotes simpler, sparser structures.

In this paper we review several ways of blending real observational data with domain knowledge conveyed by radiative transfer models. We restrict ourselves to regression problems and suggest kernel methods~\citep{Rojo18dspkm} that optimally fuse different data modalities (A1) as well as search for matching all higher-order moments of the predictive distributions (A2). An interesting form of constrained optimization includes statistical independence priors with kernels (A3), which allows us to combine observations and mechanistic models.

\subsubsection{B. Surrogate modelling and emulation} 

Emulating models in geosciences, climate sciences and remote sensing is gaining popularity \citep{camps2016survey,Reichstein19nat,camps2019perspective}. Emulators are ML models that mimic the forward physical models using a small, yet representative, dataset of simulations. Once trained, emulators can provide fast forward simulations, which in turn allows improved model inversion and parametrizations. However, replacing physical models with machine learning models require running expensive offline evaluations first, and alternatives exist that construct the model and choose the proper simulations iteratively which are summarized here \citep{camps2018physics,Svendsen19amogape}. This topic is related to active learning and Bayesian optimization, and is summarized in B1. 

\subsubsection{C. Learning and improving parameterizations} 

Physical models require setting parameters that can be seldom derived from first domain principles. Machine learning (ML) and deep learning (DL) in particular can learn such parameterizations. For example, instead of assigning vegetation parameters empirically (or sometimes even arbitrarily) to plant functional types in an Earth system model, one may allow these parameterizations to be learned from proxy covariates with ML, thus allowing some flexibility, adaptiveness, dynamics, and context-dependent properties~\citep{moreno2018methodology}. 
We briefly review here novel methods for learning parameterizations based on variational inference and Monte Carlo approaches (C1).

\subsubsection{D. Learning physics from data}

An important step towards ML models that incorporate physics are the so-called physics-informed neural networks (PINN) that directly encode nonlinear ordinary differential equations (ODE) and partial differential equations (PDE) in deep learning architectures while allowing for end-to-end training~\citep{raissi2018,raissi2019,yang2019}. Instead of using standard network layers, the authors propose a framework to directly encode nonlinear differential equations in the network that is fully end-to-end trainable. This idea allows to learn yet unknown correlations and to come up with novel research hypothesis in a data-driven way, a central point also raised in the previous research direction of interpretability. 

Here we introduce three novel alternatives. First, probabilistic models like Gaussian processes allow encoding ODEs as a form of convolutional process (D1), which report additional advantages: besides the uncertainty quantification and propagation, they also learn the explicit form of the driving force and the ODE parameters, offering a solid ground for model understanding and interpretability \citep{Svendsen20convproc}. 
These methods allow to encode a priori knowledge in the form of ODEs in machine learning to estimate hybrid parameterized models. A more direct (and challenging) approach is that of directly {\em discovering} ODEs from data (D2). In this paper we will explore the use of sparse-regression methods \citep{Brunton16} to infer the governing equations of biosphere processes from Earth observation data. A third alternative considers interpolation approximations inside Gibbs samplers (D3).

\section{In the Physics - Machine Learning Interplay}

We illustrate different ways of model-data `hybridization' in the context of Earth sciences. Modern machine learning methods that can encode differential equations from data, constrain data-driven models with physics priors and dependence constraints, improve parameterizations by variational forward-inverse modeling, emulate physical models, and blend data-driven and process-based models. This is a collective long-term AI agenda towards developing and applying algorithms capable of discovering knowledge in the Earth system.

\subsection{A1. Blending data and simulations with Joint GPs}

Biophysical parameter retrieval is often tackled with machine learning models, such as Gaussian Processes (GPs). They learn a mapping from an observed satellite spectrum to an underlying biophysical parameter. Unfortunately, real data in a remote sensing context requires expensive and time consuming terrestrial campaigns. On the other hand, there is a wealth of simulated data available, through physical models such as radiative transfer models (RTMs).   
There are obvious reasons for attempting blending real and simulated data, for instance avoiding overfitting in data scarce regimes while being consistent and allowing for extrapolation, respectively.  
This is why one would like to have the simulated data {\em guiding} a ML model when extrapolating into the region of scarce data, but doing so without confusing predictions made in the {\em data-rich} domain. 

A joint GP (JGP) modeling was introduced in \citep{svendsen17jgp} to automatically detect the relative quality of the simulated and real data, and combine them accordingly. This occurs by learning an additional hyper-parameter w.r.t. a standard GP model, and fitting parameters through maximizing the pseudo-likelihood of the real observations. If the simulated data is not helpful for prediction, the term takes on a low value which forces the weights on the simulated data to zero.
Connections of the JGP model to Hierarchical Bayesian approaches \citep{gelman2013bayesian}, and multitask focused GPs \citep{leen2011focused} can be established.

\begin{figure}[h!]
\centering
\includegraphics[width=9cm]{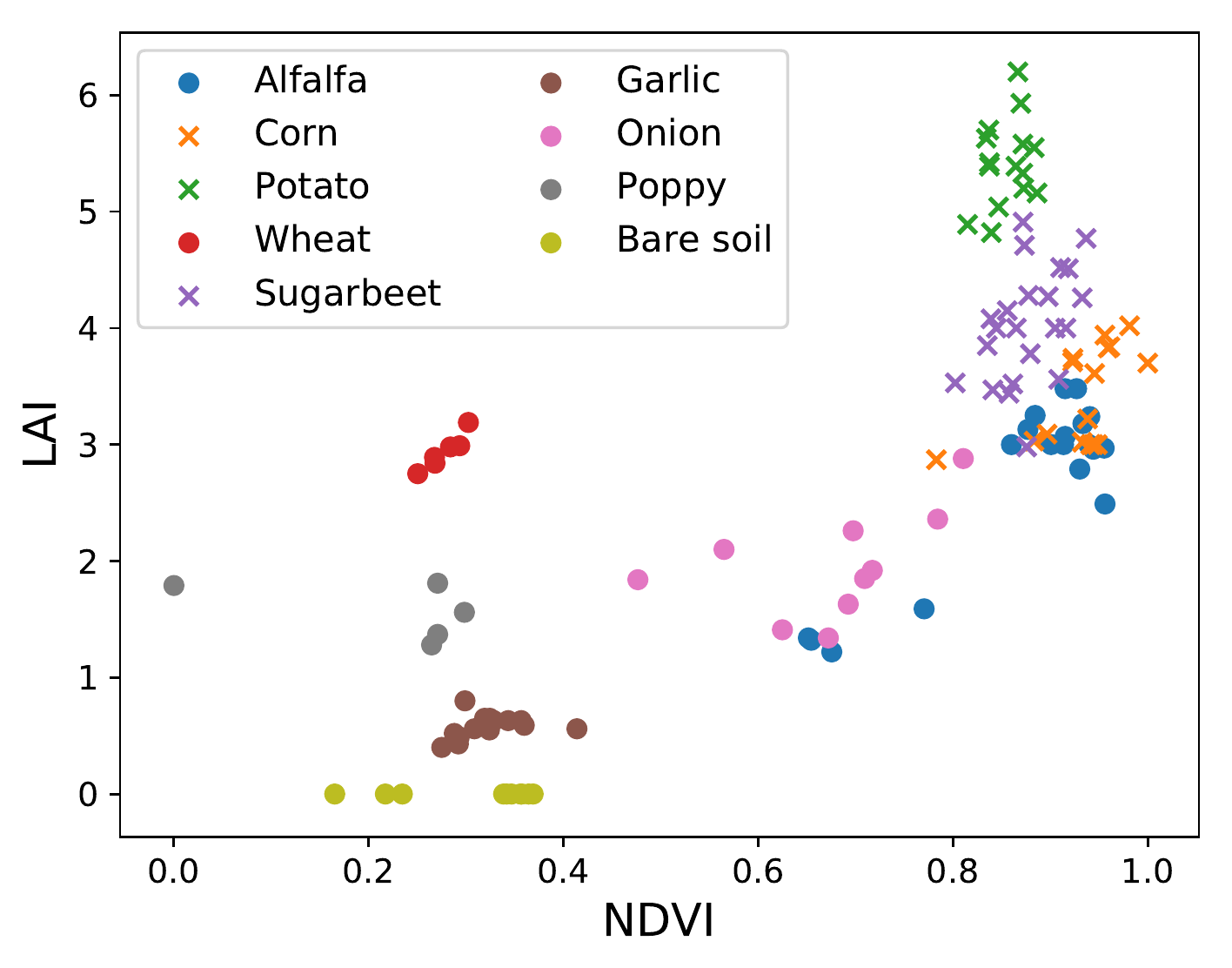}
\caption{ \label{fig:ndvi_lai} Projection of data onto the NDVI-LAI space, showing how different crop types tend to cluster together.}
\end{figure}

The JGP model helps in extrapolation or skewed sampling settings. Let us illustrate JGP performance on {the} estimation of leaf area index (LAI) from satellite imagery 
{when} the dataset is biased towards low LAI values, see Fig.~\ref{fig:ndvi_lai}. 
In this situation, a normal GP model would have a hard time extrapolating to the high-LAI values (crosses). The JGP incorporates simulated data generated through the PROSAIL RTM. Table~\ref{tab:rmse_jgp} shows that the JGP performs better than working {with} real data only or with the real+simulated data stacking. {This suggests that JGP manages} to use the simulated data for extrapolation while still using {real data information.}

\begin{table}[h!]
\centering
\caption{\label{tab:rmse_jgp} RMSE for different GP schemes.}

\begin{tabular}{l|c|c|c|c}
\hline\hline
Method & GP$_\text{R}$ & GP$_\text{S}$ & GP$_\text{R+S}$ & JGP \\ \hline
RMSE   & 1.72 & 1.64 & 1.70 & 1.60 \\
\hline\hline
\end{tabular}
\end{table}

\subsection{A2. Regression for fusion with distribution priors}

An alternative approach to data fusion may consider enforcing that not only the prediction error is reduced but also the estimates on real and simulated data show similar multivariate distributions. This requires including a new term, a prior, in the loss term that enforces predictive distributions similarity. 
\begin{multline}
    \mathcal{L} = \mu \underbrace{\|{\bf y}-f_{KRR}({\bf X};{\bf w})\|^2}_{error~fit} 
    + \lambda \underbrace{\| {\bf y}' - f_{KRR}(\hat f({\bf y}');{\bf w})\|^2}_{consistency~fit}\\
    + \nu \underbrace{\|P(f_{KRR}({\bf X};{\bf w})) -  P({\bf y})\|_{\text{DIST}}^2}_{density~fit},
    \label{eq:loss}
\end{multline}
where $\mu$, $\lambda$ and $\nu$ are regularization hyperparameters calculated performing a grid search. 
The Maximum Mean Discrepancy (MMD) measure~\citep{Rojo18dspkm} is a kernel measure of similarity between distributions which allows us to define this extra term in the standard kernel regression setting. See \citep{adsuara2019} for recent applications of distribution regression in remote sensing. Interestingly, the new kernel distributional regression shows a closed-form solution.
The results show how the use of the three constraints increase the performance of the model when the real training data is scarce and does not represent faithfully the distribution of the physically-aware data, represented here as real validation.

\begin{table}[h!]
    \centering
    \caption{Statistics from the dataset.}
    \begin{tabular}{@{}lccccc@{}}
    \hline
    \hline
              & min    & max     & mean   & std    & \#samples \\ \hline
    Real      & 0.0    & 6.5000  & 1.2549 & 1.6351 & 12261     \\
    Sim. & 0.0013 & 10.4980 & 2.9905 & 2.3037 & 14700     \\
    \hline
    \hline
    \end{tabular}
    \label{tb:tb_stats}
\end{table}

\begin{figure}[h!]
   \centering
    \includegraphics[width=6cm]{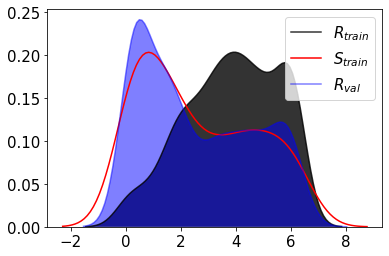}
    \includegraphics[width=6cm]{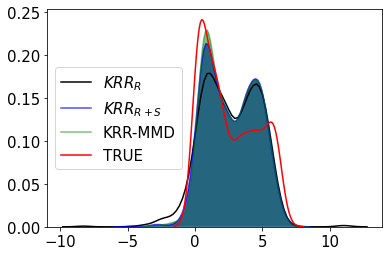}
    \caption{Distributions of (a) the original data and (b) the predicted targets by the Kernel Ridge Regression (KRR), KRR$_\text{R+S}$ and KRR-MMD.}
    \label{fig:hist}
\end{figure}

We illustrate the performance of the method in a case of fusing real satellite observations from the Moderate Resolution Imaging Spectroradiometer (MODIS) sensors (in $577$ sites over $36$ dates, i.e. $n = 20772$, ending in $12261$ points after removing the invalid values) and PROSAIL simulations (LUT containing $14700$ points) to produce LAI estimates from land surface reflectance data. Looking at data characteristics, they suggest some kind of distributional mismatch, cf. Table~\ref{tb:tb_stats}. 
It is often the case that real data {are} scarce and {show} an incomplete distribution. In our experiments, the distribution of the real dataset does not match the distribution of the testing dataset,
see Fig.~\ref{fig:hist}(a). Results in Fig.~\ref{fig:hist}(b) suggest that adding the MMD constraint helps matching the distributions of the real and simulated data, which then translates in improved accuracy scores (Cf. Table~\ref{tb:result}).

\begin{table}[h!]
    \centering
    \caption{Results for the 5-CV.}
    \begin{tabular}{lccc}
    \hline\hline
    \textbf{Model} & \multicolumn{1}{l}{\textbf{R$^2$}}        & \multicolumn{1}{l}{\textbf{RMSE}} & \multicolumn{1}{l}{\textbf{MAE}} \\ 
    \hline\hline
    KRR$_{\text{R}}$    &         0.7337  &         0.1547   &         0.1008 \\
    KRR$_{\text{R+S}}$  &         0.8058  &         0.1328   &         0.0950 \\
    \hline
    KRR-MMD             & \textbf{0.8142} & \textbf{0.1299}  & \textbf{0.0948} \\ 
    \hline\hline
    \end{tabular}
    \label{tb:result}
\end{table}

\subsection{A3. Hybrid modeling with dependence priors}

A standard family of hybrid modeling can be framed as a constrained optimization problem, where the physical rules (or model) are included as a particular form of regularizer \citep{bk:vapnik98,rn:svm,von2019informed}. This approach has been proposed before with neural networks \citep{Karpatne17}. Alternatively, here we consider a notion of algorithmic fairness in kernel methods following \citep{perez-suay_fair_2017}, which states that the learning process should be as statistically independent of variables {sensitivity} to bias as possible. Our hypothesis is that fairness and domain knowledge play similar roles: imposing physical consistency can be achieved by enforcing that model predictions {are} statistically dependent on the physical process, in the same way as being algorithmically fair should assume that model predictions {should be} independent of a sensitive covariate. The extra regularizer added to the loss function is based on the Hilbert-Schmidt Independence Criterion (HSIC) \citep{Gretton05}. The proposed method is called fair kernel learning (FKL) \citep{perez-suay_fair_2017} and enforces model predictions to be not only accurate but also consistent with a physical model, simulations, or ancillary observations.

\begin{figure}[h!]
\centerline{\includegraphics[width=8cm]{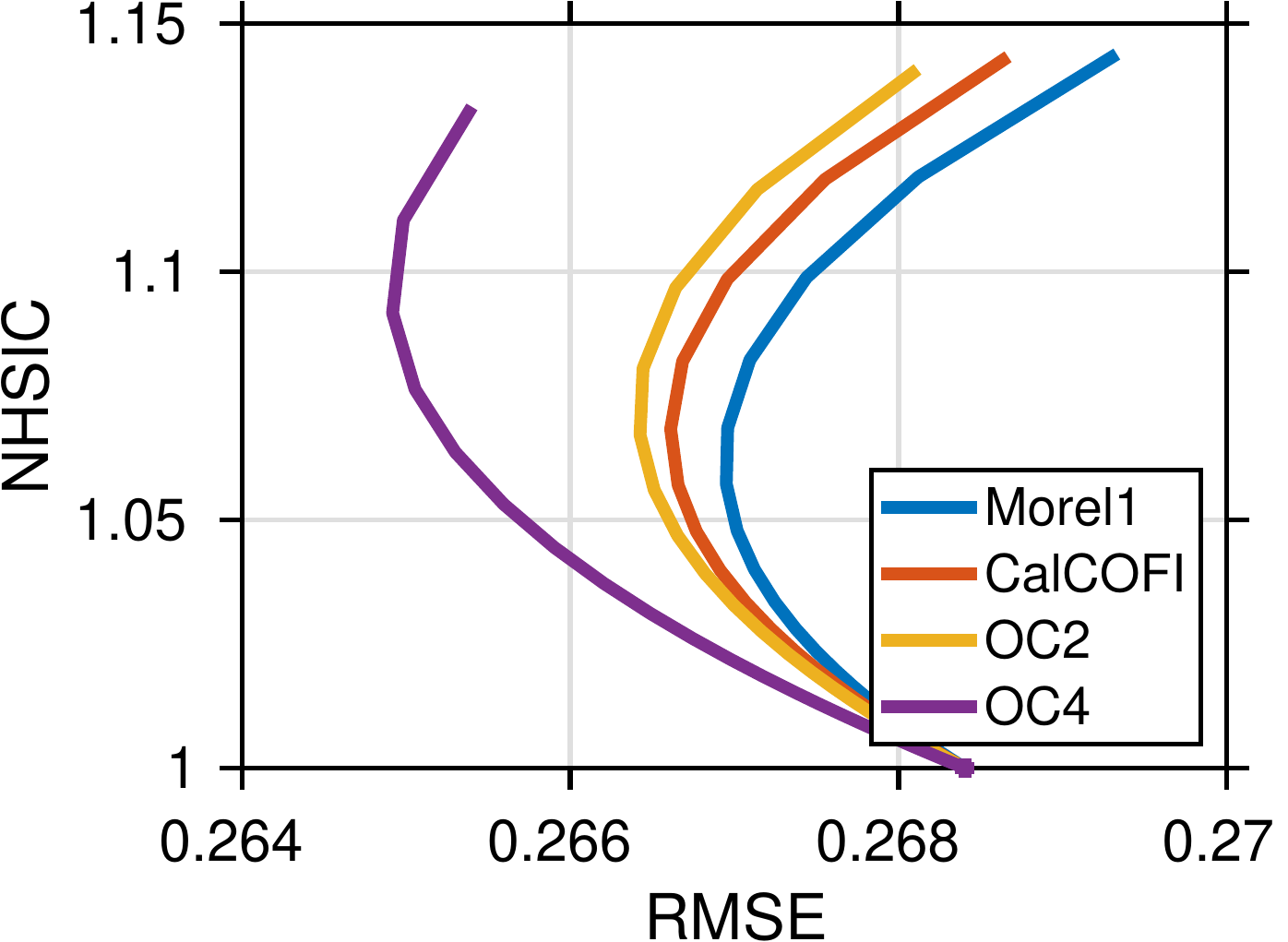}}
\caption{Consistency curves of error-vs-dependency when including parametric models into the kernel method via a normalized HSIC regularization term.}  \label{fig:phys_Chlorophyll}
\end{figure}

As an illustrative example, let us assess the performance of the FKL model for the estimation of chlorophyll concentration from multispectral images. We used the SeaBAM dataset \citep{OReilly98}, which gathers 919 {\em in-situ} measurements of chlorophyll concentration around the United States and Europe measured by the SeaWiFS ocean color satellite sensor. We forced dependence with respect to four standard ocean color parametric models (Morel1, CalCOFI 2-band linear, OC2 and OC4) in the model, and trained it to estimate ocean chlorophyll content $y$ from input radiances $x$.  Results in Fig. \ref{fig:phys_Chlorophyll} show that including the regularizer helps {to} reduce the RMSE {in the predictions}. The FKL also allows us to discriminate which parametric model resembles more the observational data as all models improve the accuracy but differ in the level of improvement: OC2 and OC4, lead to higher accuracy meaning that they help FKL to explain the data better.

\subsection{B1. Optimal emulation of radiative transfer models}

The introduction of emulators have captured the attention of many researchers dealing with computationally heavy models in many areas \citep{antoulas2005approximation,frangos2010surrogate}, and in remote sensing in particular with the wide use and adoption of RTMs. Emulators are essentially machine learning (regression) algorithms that provide fast approximations to complex physical (radiative transfer) models, an approach with a long story in statistics \citep{OHagan78}. These surrogate models or metamodels are generally orders of magnitude faster than the original model, and can then be used \emph{in lieu} of it, opening the door to more advanced biophysical parameter estimation methods. Training an emulator, however, requires running a number of simulations which in many cases are too costly or do not represent the space well. 
This is why selecting the optimal, more informative simulations is key to construct a look-up-table and hence the emulator. 
In \citep{Svendsen19amogape} we introduced a GP-based emulator such that a minimal number of simulations is run to attain a given approximation error. The method is related to Bayesian optimization and active learning techniques. 
The Automatic Multioutput Gaussian Process Emulator (AMOGAPE) iteratively runs a GP interpolator from which one derives an \textit{acquisition function} that describes the optimal regions to sample from in geometric and uncertainty terms. In each iteration the RTM generates the queried simulations that are included in the training dataset to update the emulator. 

\begin{figure}[h!]
\centerline{ \includegraphics[width=9cm]{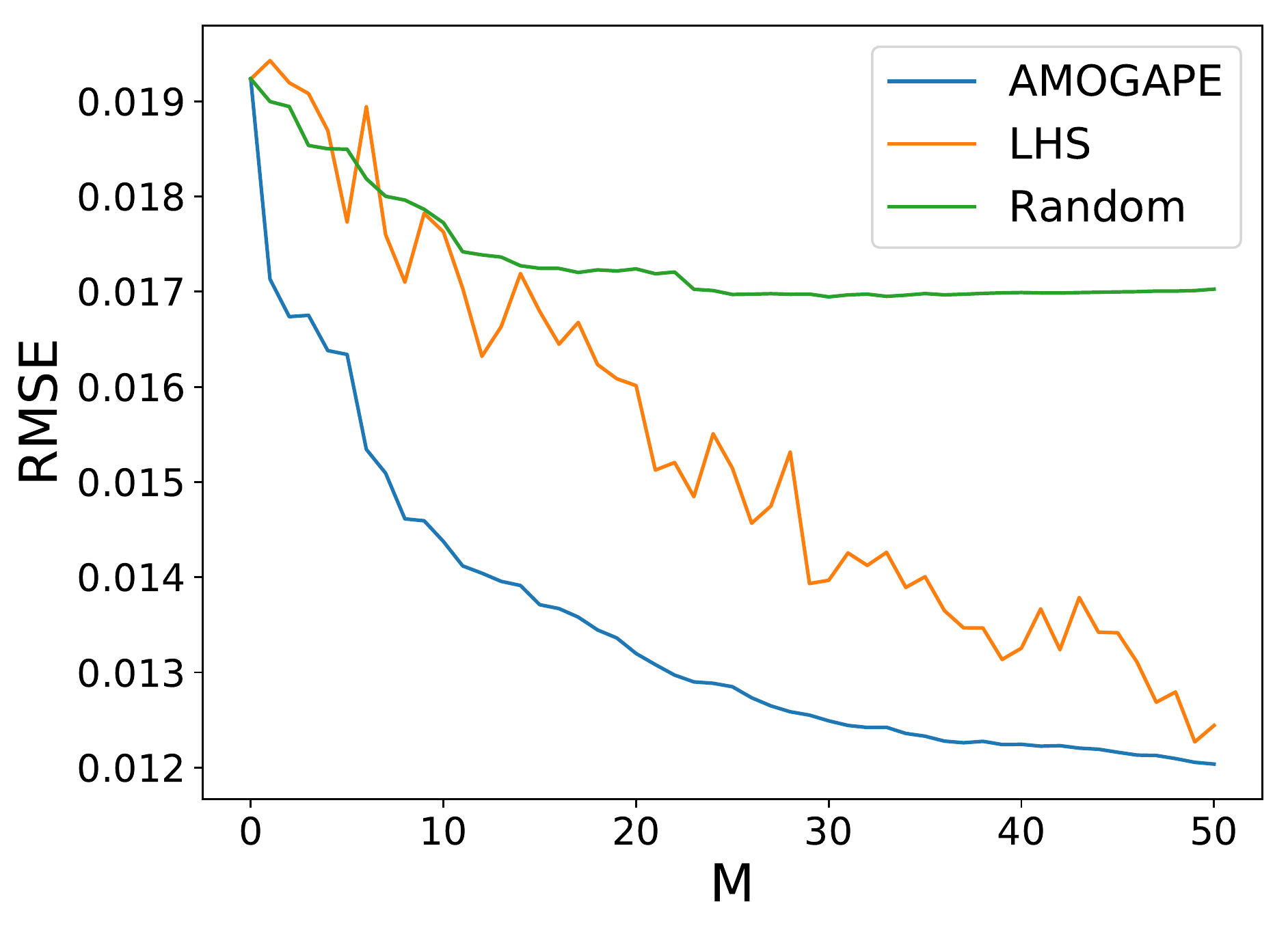} }
\caption{RMSE for emulators using random sampling, Latin Hypercube Sampling (LHS) and AMOGAPE. The loop is iterated until the stop condition is fulfilled.}
\label{fig:rmsefinal} 
\end{figure}

We tested AMOGAPE for emulation of the standard leaf-canopy PROSAIL model \citep{Jacquemoud2009b}. PROSAIL models canopy reflectance using the turbid medium assumption and is thus particularly well suited for homogeneous canopies. PROSAIL simulates leaf reflectance from 400 to 2500 nm with a 1 nm spectral resolution as a function of biochemistry and structure of the canopy, its leaves, the background soil reflectance and the sun-sensor geometry. Leaf optical properties are given by the mesophyll structural parameter ($N$) and leaf chlorophyll ($Chl$), dry matter ($Cm$), water ($Cw$), carotenoid ($Car$) and brown pigment ($Cbr$) contents. At canopy level PROSAIL is characterized by leaf area index ($LAI$), the average leaf angle inclination ($ALA$) and the hot-spot parameter ($Hotspot$). The system geometry is described by the solar zenith angle ($\theta_s$), view zenith angle ($\theta_\nu$), and the relative azimuth angle between both angles ($\Delta \Theta$). 
Among the great many leaf and canopy parameters in PROSAIL, we chose only chlorophyll and LAI, and keep the rest fixed for illustration purposes. We evaluated the RTM at all the possible 4900 combinations of $Chl$ and $LAI$.  Results were averaged over $50$ runs for several standard methods, see Fig. \ref{fig:rmsefinal}. 
The Latin hypercube sampling (LHS) method exhibits a lot of variance as it chooses a completely new set of points for each iteration. AMOGAPE identifies the needed points rather quickly, while the LHS decreases the error more slowly. The random sampling method, not being designed to maximize information gained, does not manage to reach the same level of error as the other methods. This gap is expected to widen as the input dimensionality grows.

\subsection{C1. Approximation of physical parameter densities}
There is a strong focus in the remote sensing literature on producing \textit{point estimates} of the biophysical parameters, or causes, that give rise to  satellite observations, or effects. Equally important, however, is the ability to infer the underlying probability distribution of biophysical parameters in a given area. This can be used to characterize regions and lead to better simulations.

{\bf Problem Statement.}
Let us define  the vector of effects ${\bf e} \in\mathbb{R}^{D_e}$ and vector of causes ${\bf c} \in\mathcal{C}\subseteq\mathbb{R}^{D_c}$. An RTM model, denoted ${\bf f}({\bf c}): \mathbb{R}^{D_c} \rightarrow \mathbb{R}^{D_e}$, represents the underlying link from ${\bf c}$ to ${\bf e}$.
The complete observation model is given by
\begin{eqnarray}\label{eq:generative}
    {\bf e}={\bf f}({\bf c}) + {\bm \epsilon},  \quad {\bm \epsilon} \sim \mathcal{N}({\bm \epsilon}|{\bf 0},\sigma^2\mathbf{I}),
\end{eqnarray}
where $\mathbf{I}$ is a unit $D_e \times D_e$ matrix. The observation model induces then the following likelihood function,
\begin{equation}\label{eq:llhood}
p({\bf e}|{\bf c})=\mathcal{N}({{\bf e}}|{\bf f}({\bf c}),\sigma^2\mathbf{I}).
\end{equation}
Note that by fixing ${\bf c}$, the conditional probability $p({\bf e}|{\bf c})$ is Gaussian, but as a function of ${\bf c}$ the likelihood is a highly non-linear function due to the dependence on the RTM with the causes, i.e. ${\bf f}({\bf c})$. Note that, we have access several observed effects ${\bf e}$
 We assume that some set of data ${\bf e}$ is given and an estimation of $p({\bf e})$ is possible. Therefore, we assume to be able to evaluate $p({\bf e}|{\bf c})$, $p({\bf e})$, and we are interested in approximating the other marginal $p({\bf c})$ (and also $p({\bf c}|{\bf e})$ if possible).

 \begin{figure}[h!]
    \centering
    \includegraphics[width=12cm]{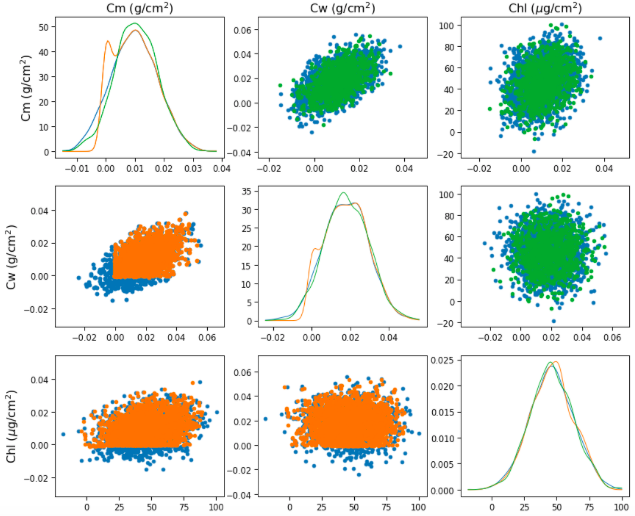}
    \caption{The blue points are ${\bf c}$'s from the training set, while the green points are drawn from the fitted prior. The diagonal shows density estimation of ${\bf c}$ using samples from the ground truth prior (blue), and the posterior approximation (orange) and the fitted prior (green).}
    \label{fig:3D_overview}
\end{figure}

{\bf Possible solution.} 
We assume an Gaussian prior over ${\bf c}$'s,  
\begin{eqnarray}
p({\bf c}) = \mathcal{N}({\bf c}|\mathbf{m}, \mathbf{S})\,,
\end{eqnarray}
where ${\bf m}\in \mathbb{R}^{D_c}$ and the $D_c\times D_c$ covariance matrix ${\bf S}$ are considered unknown.
The posterior density given the observed data ${\bf e}$ over the causes can be  expressed as 
\begin{eqnarray*}
   p({\bf c}|{\bf e})  &\propto& p({\bf e}|{\bf c})p({\bf c})= \mathcal{N}({{\bf e}}|{\bf f}({\bf c}),\sigma^2\mathbf{I})\mathcal{N}({\bf c}|\mathbf{m}, \mathbf{S}).
\end{eqnarray*}
We can learn the prior parameters, vector ${\bf m}$ and matrix ${\bf S}$, and obtain an approximation of the posterior $p({\bf c}|{\bf e})$,  which represents an inverse probabilistic mapping from ${\bf e}$ to ${\bf c}$. This can be done  using a Variational Inference (VI) scheme \citep{kingma2013auto} or by a Monte Carlo approach \citep{wei1990monte}. An example of application of VI to 3 PROSAIL variables, leaf chlorophyll (Chl), dry matter (Cm), water (Cw), is given in Fig. \ref{fig:3D_overview}.

\subsection{D.1. Learning forcings with convolution processes}

Gaussian process convolution models \citep{ver1998constructing} help to incorporate prior knowledge in a Gaussian process (GP) model. A type of process convolution named {\em latent force model} (LFM) was originally introduced in~\citep{alvarez2009latent}, and performs a convolution between underlying (latent) GP functions and a smoothing kernel derived from the ODE assumed to govern the system. 
The LFM performs multioutput regression, adapts to the signal characteristics, is able to cope with missing data in the time series, and provides explicit latent functions that allow system analysis and evaluation \citep{Svendsen20convproc}. 

\begin{figure}[h!]
    \centering
    \includegraphics[width=12cm]{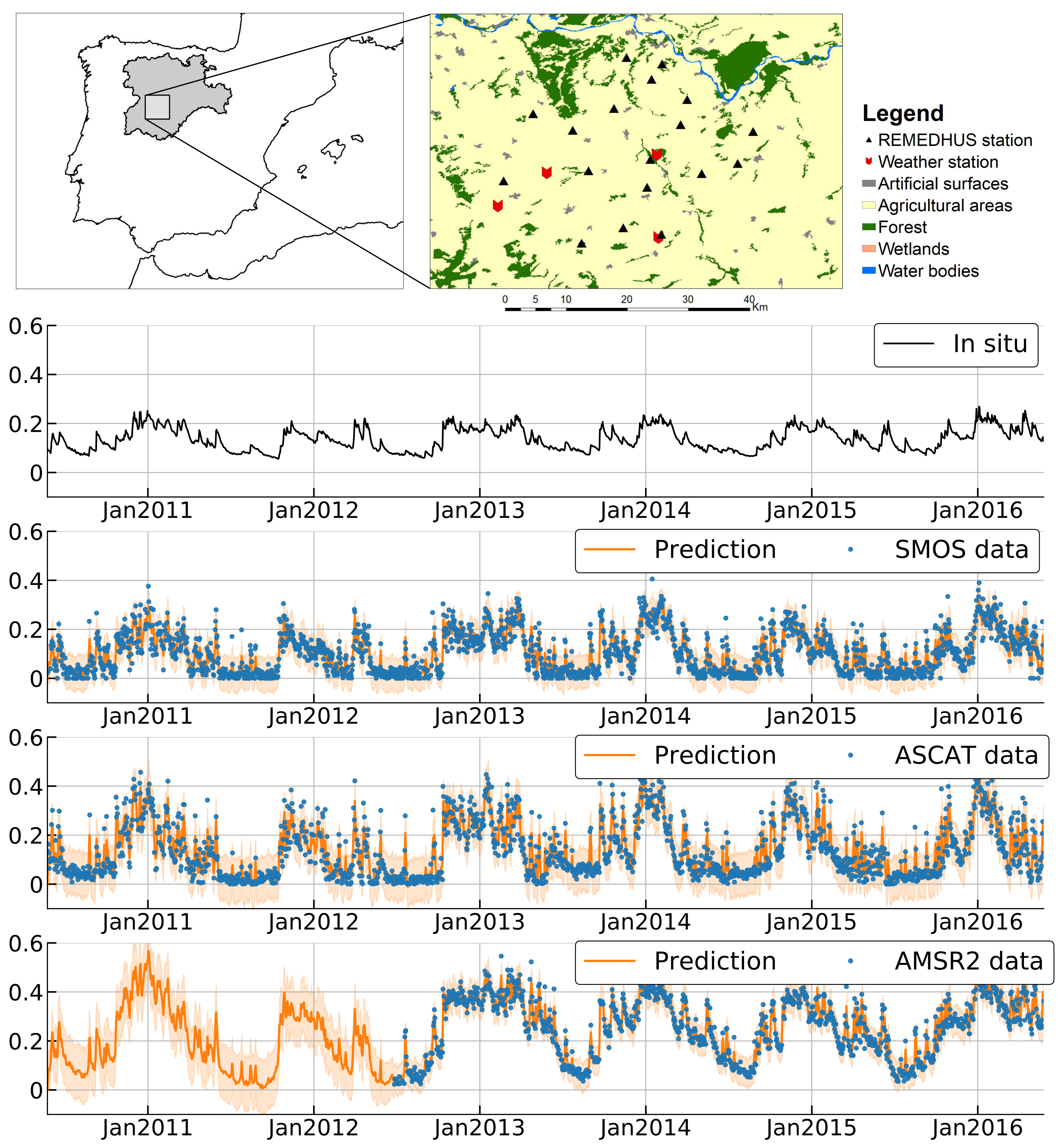}
    \caption{Multi-output process convolution results in gap filling multisensor soil moisture time series. Top: layout of in-situ stations at the REMEDHUS network \citep{Sanchez12}. Bottom: time series of in-situ (average of 18 stations) and satellite-based soil moisture estimates (m$^3\cdot$m$^{-3}$). Blue dots denote training data, purple lines and shaded regions represent LFM predictions and their confidence intervals.} \label{fig:lfmpred}
\end{figure}

We encoded in the LFM a first order differential equation modeling soil moisture (SM). This is a widely used model for this physical system describing the exponential decay in time of the water in a soil volume. We considered six years of SM estimates from three satellite microwave products: SMOS, ASCAT and AMSR2. LFM learns the covariances not only between function values of the same signal, but also between signals, which makes it extremely useful for gap-filling \citep{Svendsen20convproc}, see Fig. \ref{fig:lfmpred}. 
The reconstructed SM time series closely follow the original ones, capturing the wetting-up and drying-down events and filling in the missing information. Predictions have associated confidence intervals related to data uncertainty as well as to model uncertainty. LFM reconstructs the first two years (pre-launch) of AMSR2 by seamlessly transferring the information from the other sensors. 
More interesting is the fact that the model automatically retrieves the estimated input noise $\sigma$ and e-folding time $\tau$ (days) obtained per satellite, as well as latent forces that resemble and correlate well with the rainfall, which was never seen by the algorithm \citep{Svendsen20convproc}. 

\subsection{D2. Learning ODEs with sparse regression} 

As in many fields of science and engineering, Earth system models describe processes with a set of differential equations encoding our prior belief about the dynamics and variable interactions. 
Learning ODEs from stochastic variables is a challenging problem, which has been a subject of intense research recently: equation-free modeling \citep{ye2015equation}, empirical dynamic modeling \citep{sugihara2012detecting,ye2015equation}, modeling emergent behavior \citep{schmidt2011automated}, and automated inference of dynamics \citep{daniels2015automated,daniels2015efficient} 
are some examples. 
Other more recent approaches have considered deep learning to learn the governing equations \citep{lu2019deepxde,huang2020learning}, but they come at expensive cost. A simpler more efficient method considers least squares regression to explore the space of simplest ODEs solutions \citep{Brunton16}. 
In many of these methods, however, one makes a sufficiency assumption: only a set of descriptors, variables, class of models, interactions and (sometimes even) permitted trajectories are assumed necessary to describe the system. This modeling process is implicitly assuming a {\em sparse} class of model solutions, that is a reduced number of governing equations is permitted to model system dynamics. Many modules and submodules in radiative transfer models, climate models, and biogeochemical models are based on differential equation parameterizations encoding variable relations and dynamic behaviour. 
Following the previous rationale, we explore the use of machine learning techniques to bring systems of ordinary differential equations (ODEs) to light purely from data. 
Applying Occam's razor {logic}, we exploit sparse regularization to identify the most expressive, resolved, and simplest ODEs explaining the data in the framework of \citep{Brunton16}. 

\begin{figure}[h!]
    \centering
    \includegraphics[width=6.1cm]{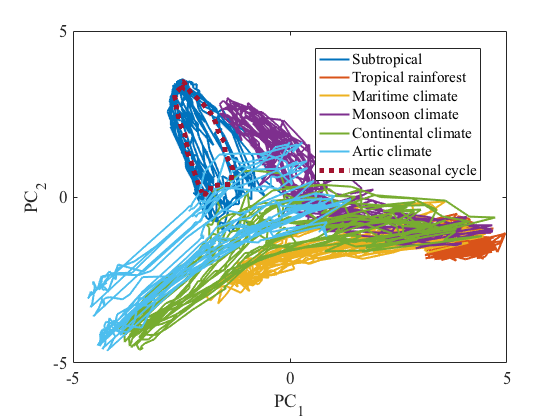}
    \includegraphics[width=6.1cm]{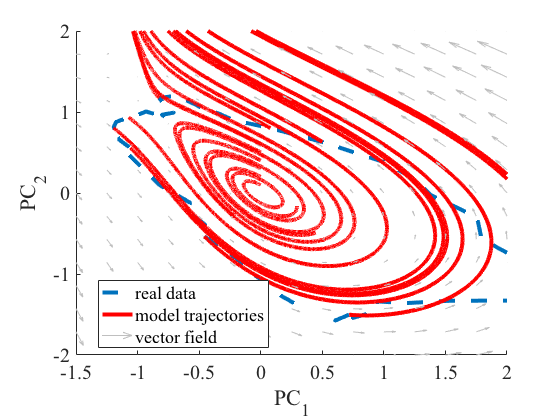}
    \caption{(a) Trajectories in the phase space of the first two principal components for representative ecosystems. (b) Phase space with the vector field corresponding to the analytic model discovered (gray), possible trajectories from different initial conditions (red), and the real data  (blue)}.
    \label{fig:mscMexico}
\end{figure}

We illustrate the methodology to describe 12 biosphere indices proposed in \citep{Kraemer19} for describing land surface dynamics. 
Fig.~\ref{fig:mscMexico}(a) shows trajectories in the phase space of the first two principal components for some ecosystems. The sparse model yielded high accuracy ($R=0.75$) and a set of ODEs:
\begin{equation}
\begin{array}{rcl}
\frac{d}{dt}\text{PC}_1 &=& - 37.5 \text{PC}_1 - 55.6 \text{PC}_2 - 31.9 \text{PC}_1\text{PC}_2 \\
\frac{d}{dt}\text{PC}_2 &=& + 67.2 \text{PC}_1 + 44.8 \text{PC}_2 - 74.0 \text{PC}_1\text{PC}_2, \nonumber
\end{array}
\end{equation}
where the PC$_1$ (mainly describing productivity) decreases exponentially if PC$_2$ (accounting for water availability) were not available (as the second and third terms in the first ODE would vanish). The cross-term actually just states that productivity increases with water availability (note that in this case $\text{PC}_1<0$ while $\text{PC}_2>0$). The opposite behaviour is observed for the second ODE; increase in water availability is only possible with a decrease in productivity (due to vegetation loss during the winter period, for example).
Fig.~\ref{fig:mscMexico}(b) shows the retrieved phase space, as well as some possible trajectories from different initial conditions, for the particular case of the subtropical ecosystem corresponding to the area of Mexico. 

\subsection{D3. Gibbs for complex system and ODE learning}

\begin{figure}[htb]
\centering 
\centerline{
 \includegraphics[width=6.1cm]{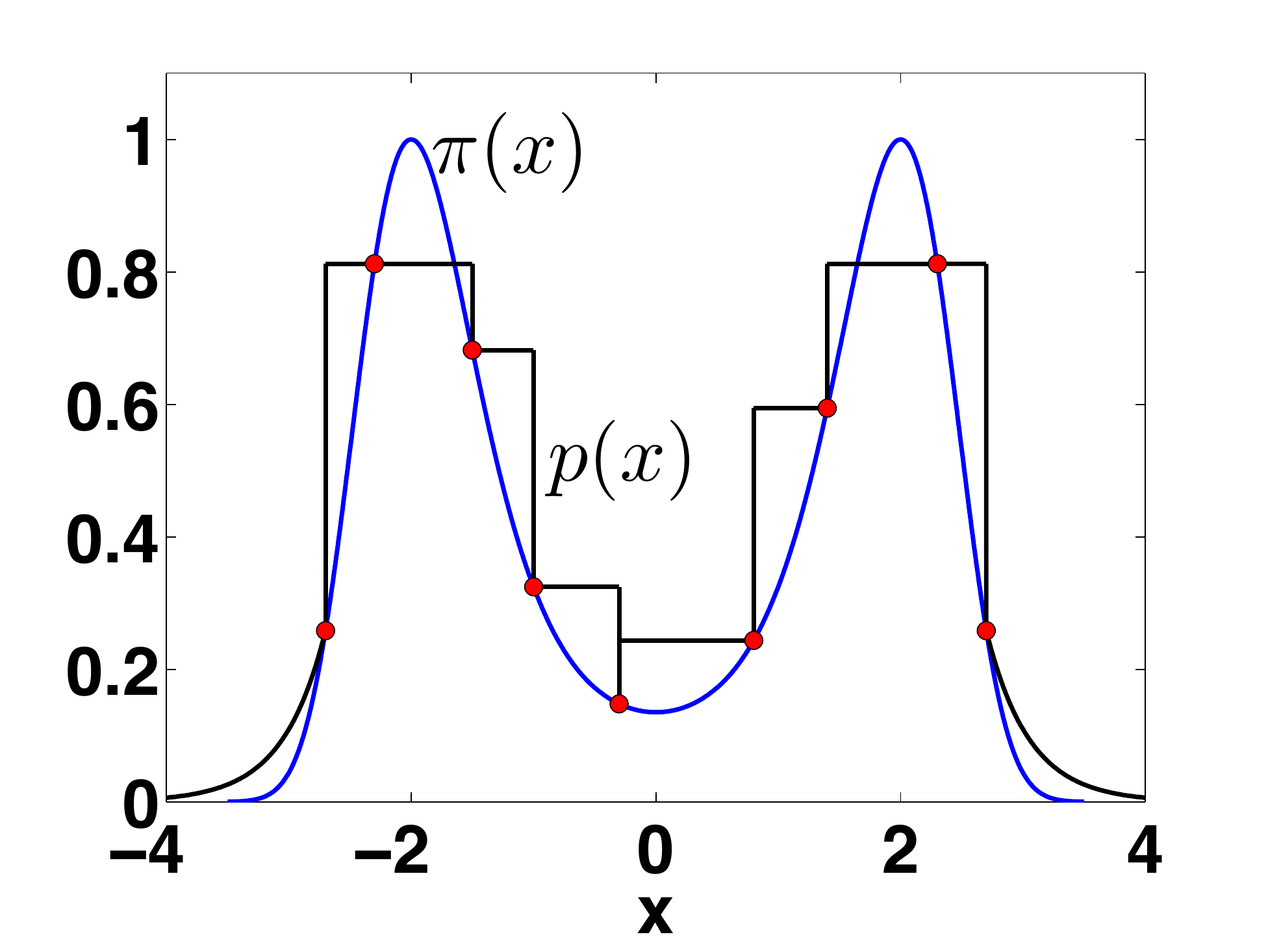}
 \includegraphics[width=6.1cm]{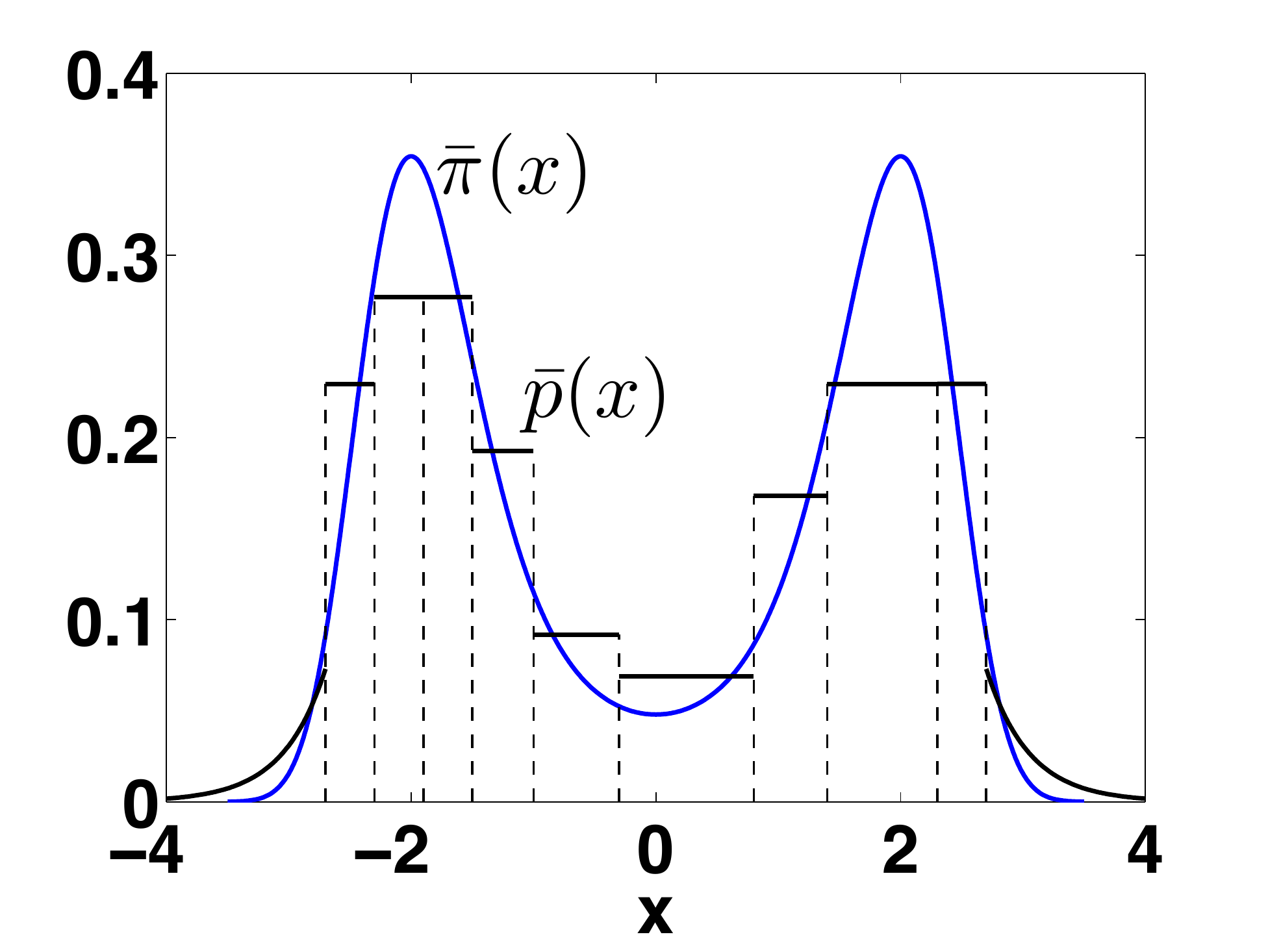}
}
  \caption{Example of the proposal construction $p(x)$ for a specific full-conditional, using an interpolation procedure (left) and then normalizing it, obtaining  $\bar{p}(x)$ (right).}
\label{figFUSSconst}
\end{figure}

Let us consider the problem of inferring  a vector of parameters, denoted as ${\bf x}$, of a complex system, such as the parameters of a  ODE or a discrete difference equation, Let us also denote the vector of observed data as ${\bf y}$, obtained as output of the complex mathematical system.
Bayesian inference often requires drawing samples from complicated multivariate posterior pdfs, $\pi({\bf x}|{\bf y})$ with 
$$
{\bf x}=[x_1,...,x_D] \in \mathcal{X}^D \subseteq \mathbb{R}^D.
$$
A common approach, when direct sampling from $\pi({\bf x}|{\bf y})$ is unfeasible, is using a Gibbs sampler \citep{MARTINO201568}.
At the $i$-th iteration, a Gibbs sampler obtains the $d$-th component ($d=1, \ldots, D$) of ${\bf x}$, $x_d$, by drawing from the full conditional pdf of $x_d$ given all the previously generated components, i.e.,
\begin{align}
	x_d^{(i)} \sim&
	\bar{\pi}(x_d |x_{1}^{(i)},x_{2}^{(i)},....x_{d-1}^{(i)},  x_{d+1}^{(i-1)},..., x_{D}^{(i-1)}) \\
	&= \bar{\pi}_d(x_d) \propto \pi_d(x_d),
\label{eq:gibbs}
\end{align}
However, even sampling from the univariate pdfs in Eq. \eqref{eq:gibbs} can often be complicated. In these cases, a common approach is to use another Monte Carlo technique (e.g., rejection sampling (RS) or the Metropolis-Hastings (MH) algorithm) within the Gibbs sampler, drawing candidates from a simpler proposal pdf, $\bar{p}(x_d) \propto p(x_d)$. The FUSS ({\it ''A fast universal self-tuned sampler''}) algorithm considers a piecewise interpolative approximation of $\bar{\pi}_d(x_d)$ for building $\bar{p}(x_d)$ (see Fig. \eqref{figFUSSconst}). This ensures a robust and efficient Gibbs sampler.

\begin{figure}[!ht]
  \centering
  \centerline{
    \includegraphics[width=6cm]{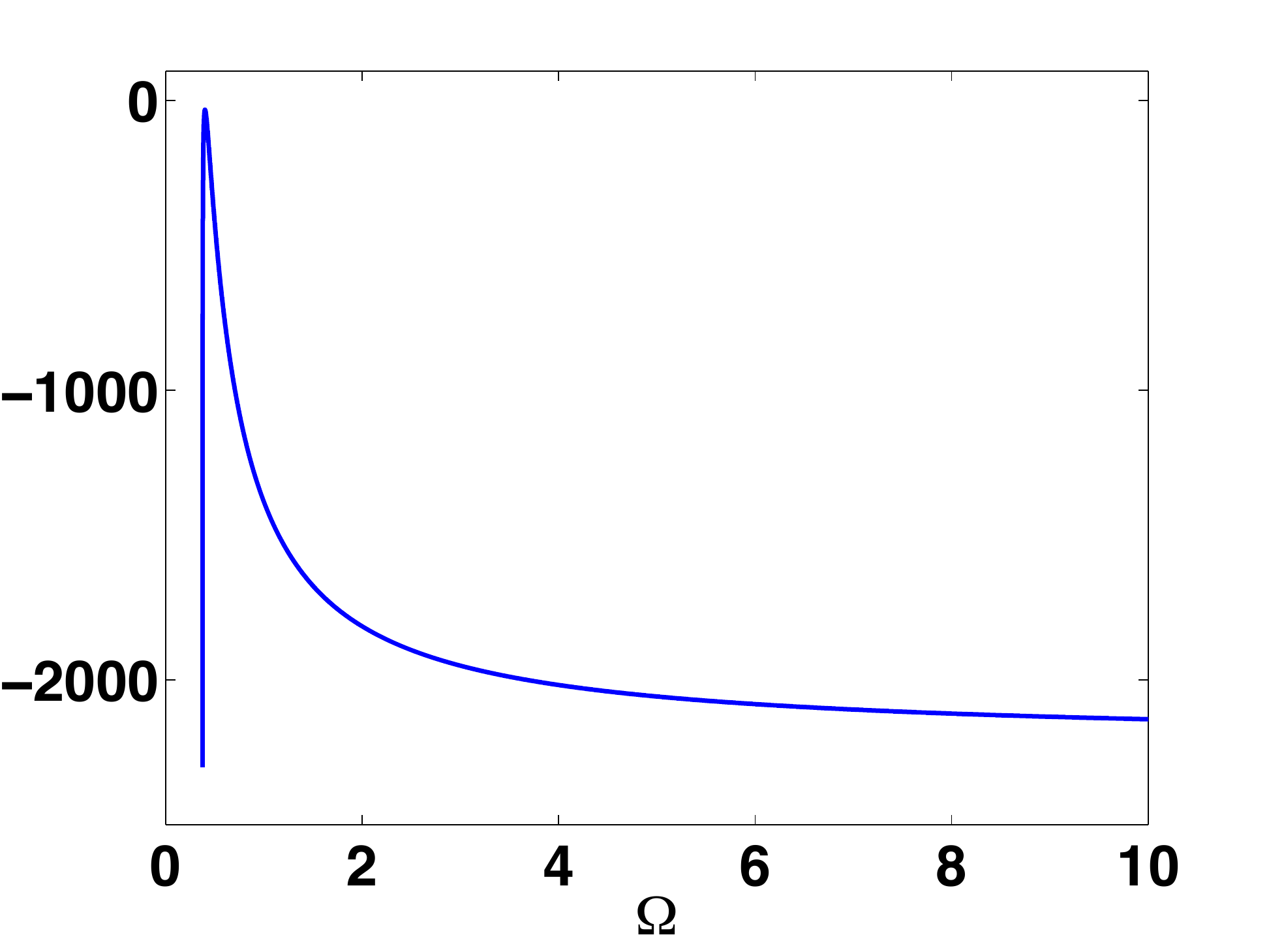}
    \hspace{-0.2cm}
    \includegraphics[width=6cm]{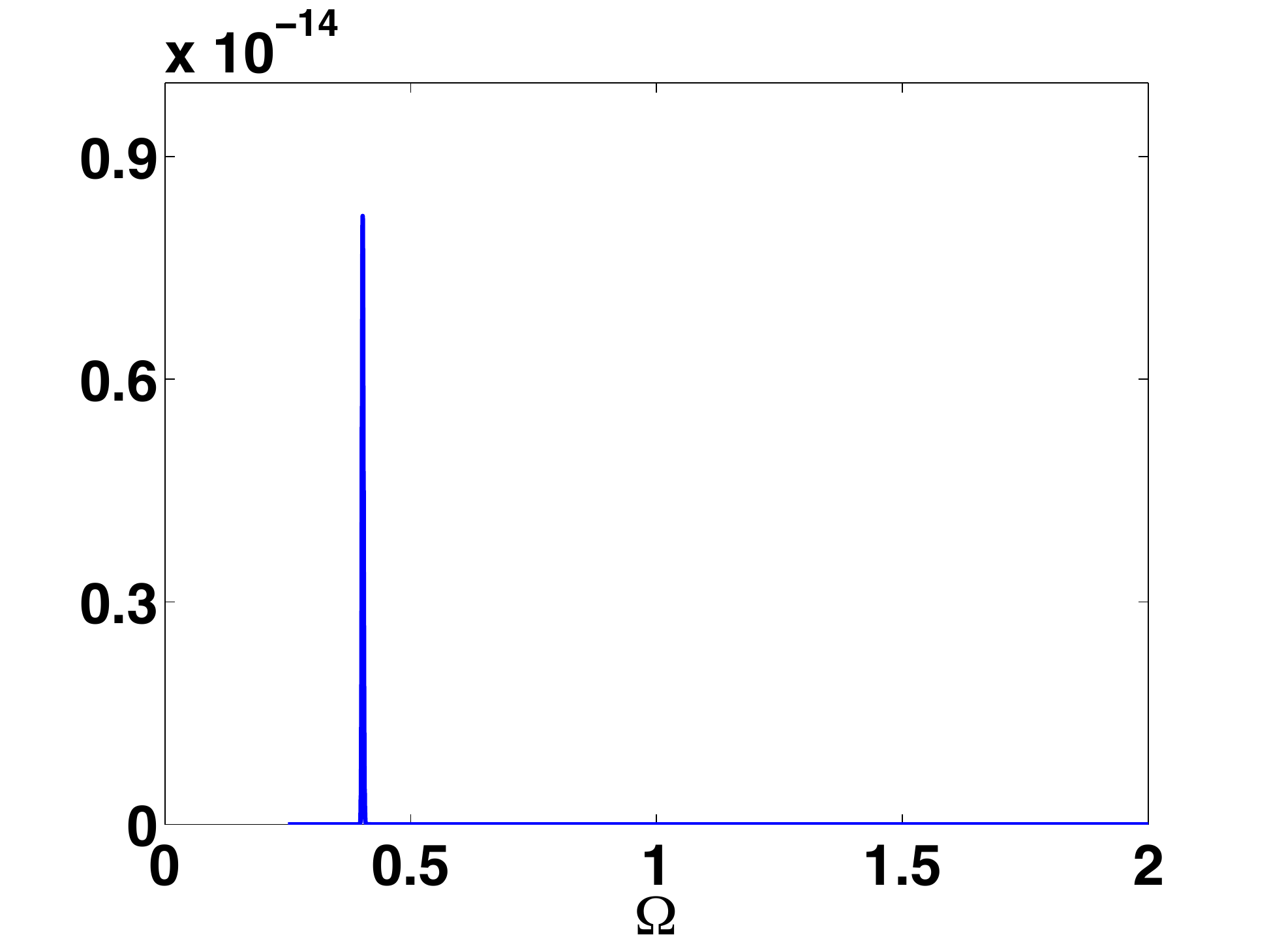}
  }
\caption{Examples of log-conditional posterior and conditional posterior  of $\Omega$ (given $R=4$).}
\label{fig:Perretti}
\end{figure}

The FUSS-within-Gibbs has been applied with success for estimating parameters in a chaotic system, which is considered a very challenging problem in the literature.
Let us consider a logistic map perturbed by multiplicative noise,
\begin{equation}
	y_{t+1}=R\left[\ y_{t}\left(1-\frac{y_t}{\Omega}\right)\right]\exp(\epsilon_t), \quad  \epsilon_t \sim \mathcal{N}(0,\lambda^2), 
\label{LogNoisy}
\end{equation}
with $y_1 \sim \mathcal{U}([0,1])$ and for some unknown parameters $R>0$ and $\Omega>0$ (we set$\lambda=0.01$).
The likelihood function is given by
$p(y_{1:T}|R,\Omega)=\prod_{t=1}^{T-1} p(y_{t+1}|y_t,R,\Omega)$. Considering uniform priors over the parameters, the FUSS-with-Gibbs obtains a MSE of order $10^{-4}$ in the estimation of $\Omega $ and $R$, compared with a MSE of $\approx 0.65$ with a standard MH-within-Gibbs. The cause of these results are the complex and sharp full-conditionals (see Fig. \ref{fig:Perretti}).

\section{Discussion and Conclusions}

We presented several ways to live in the Physics and machine learning interplay in the discipline of Earth observation. We introduced novel machine learning methods that can encode differential equations from data, constrain data-driven models with physics priors, kernel-based distribution matching and dependence constraints, improve parameterizations by variational forward-inverse modeling, emulate physical models, and blend data-driven and process-based models. 
The hybrid AI agenda introduced here in the context of Earth sciences is aimed towards developing and applying algorithms capable of discovering knowledge in the Earth system. Algorithms can be of help in other scientific disciplines and for other problems involving models and data, forward and inverse modeling, and in cases where domain knowledge and observational data are available. 

We believe that encoding domain knowledge in machine learning may not only achieve improved performance, generalization and extrapolation capabilities but, more importantly, may lead to improved consistency and credibility of the machine learning models. As a byproduct, the hybridization has an interesting regularization effect, given that the physics limits the search parameter space and thus discards implausible models. Therefore, physics-aware machine learning models combat overfitting better, become simpler (sparser), and require less amount of training data to achieve similar performance.
Explainable AI, abstraction, fairness and causal reasoning which share similar aims and connections are on their way: achieving explanatory models, understanding through data, observations, models and assumptions. 
Overall, the hybrid modeling framework constitutes a new research direction that deserves to be pursued further and with more intensity. 

\section{Acknowledgments}

Research funded by the European Research Council (ERC) under the ERC-CoG-2014 SEDAL project.


\end{document}